\documentclass[journal,10pt]{IEEEtran}
\usepackage{amsmath,amsfonts,amssymb}
\usepackage{amsthm}
\usepackage{algorithmic}
\usepackage{algorithm}
\usepackage{array}
\usepackage{textcomp}
\usepackage{stfloats}
\usepackage{url}
\usepackage{verbatim}
\usepackage{color}
\usepackage{graphicx}
\usepackage{cite}
\usepackage{subcaption}
\usepackage{booktabs}
\usepackage{float}
\usepackage{nth}
\usepackage{bm,flushend}

\allowdisplaybreaks

\newtheorem{remark}{Remark}
\begin{document}

\title{Learning to Distributedly Estimate under Partially Known Dynamics:
A Covariance-Agnostic Neural Kalman Consensus Filter}

\author{George Stamatelis,~\IEEEmembership{Student Member,~IEEE}, Kyriakos Stylianopoulos,~\IEEEmembership{Student Member,~IEEE},\\ and George C. Alexandropoulos,~\IEEEmembership{Senior Member,~IEEE}
\thanks{The authors are with the Department of Informatics and Telecommunications, National and Kapodistrian University of Athens, Panepistimiopolis Ilissia, 16122 Athens, Greece. 
    (e-mails: \{georgestamat, kstylianop alexandg\}@di.uoa.gr).}
    \thanks{
    This work has been supported by the SNS JU project 6G-DISAC under the European Union's Horizon Europe research and innovation programme under Grant Agreement No 101139130. G. Stamatelis was supported by the Hellenic Foundation for Research and Innovation (HFRI) under the 5th Call for HFRI PhD Fellowships (Fellowship Number: 21080).
}
}



\maketitle

\begin{abstract}
Online latent state estimation constitutes a fundamental challenge within the artificial intelligence field, serving as a foundational tool for diverse applications, including sequential decision making, anomaly and change-point detection. In this paper, a novel online distributed sensing framework, where agents collaborate and exchange information to perform latent state estimation, is presented. The proposed estimator combines available partial domain knowledge with the representation capabilities of deep neural networks. In particular, the designed sensing framework incorporates prior estimates, optimized consensus weights, and Kalman-like recursive updates to perform decentralized inference, without relying on knowledge of noise statistics. Extensive experiments on linear, chaotic (Lorenz), and practical wireless tracking environments reveal that the proposed Covariance-Agnostic Neural Kalman Consensus Filter (CA-NKCF) outperforms traditional distributed Kalman and particle filters as well as purely model-free deep neural networks, exhibiting robustness even when the underlying motion and observation models are misspecified. It is also demonstrated that CA-NKCF's performance advantage remains stable across varying noise levels, random communication topologies, latent state dimensions, and observation clutter densities induced by scattering objects in wireless systems.
\end{abstract}

\begin{IEEEkeywords}
Distributed estimation, tracking, Kalman filter, consensus, model-based deep learning.
\end{IEEEkeywords}

\section{Introduction}
Decentralized decision making under uncertainty is a cornerstone of Multi-Agent (MA) Artificial Intelligence (AI)\textcolor{black}{~\cite{lumimas,imas}}. A key prerequisite for efficient decentralized decisions is accurate  online estimation (a.k.a. \textit{filtering}) of latent state representations. 
One possible application is in robotics, where a group of robots explore their surrounding space, process observations using different sensors, and exchange messages in order to track moving targets; the latent state could be the position or velocity of targets~\cite{EKF-ref1,EKF-ref2,KFApplications}. Another possible application is in large-scale sensor networks tasked with monitoring a physical phenomenon, e.g., spreading fire or a gas leak\textcolor{black}{~\cite{distributedUAVWildFire,distributedGasLeak}}. Due to bandwidth limitations, sending all sensing data to a central node, access point, or Base Station (BS) may be infeasible or too time consuming for safety applications. Alternatively, each sensor may exchange local readings with nearby peers to track the underlying system state. 

In the centralized, single-sensor domain, the most established estimator is the Kalman Filter (KF)~\cite{KFOriginal} and its extensions, such as the Extended (EKF)~\cite{EKF-ref1} and Unscented (UKF) KF~\cite{UKF}. This filter is the optimal estimator for linear systems with zero-mean Gaussian noise. In fact, despite being developed over 50 years ago, it finds numerous engineering applications as well as emerging AI applications~\cite{KF_actorCritic,KFMultiAGentSystems}. However, for  realistic nonlinear systems, KF extensions are suboptimal. Furthermore, one key limitation of such Model-Based (MB) approaches is that they require precise specification of the underlying system dynamics. This feature actually renders them unsuitable in systems where the state and observation generating functions are approximations of the true dynamics. Additionally, in some high-dimensional systems, the noise can be complicated, nonstationary and, hence, it is infeasible to accurately model it~\cite{AI_KF-survey}. 

On the other hand, purely Model-Free (MF) approaches, such as end-to-end Recurrent Neural Networks (RNNs)~\cite{LSTM,GRU}, do not utilize any domain knowledge, usually sacrificing performance. To that end, KFs aided by artificial Neural Networks (NNs)~\cite{AI_KF-survey} have emerged as a powerful synergy, combining the representational power of deep NNs with the available domain knowledge provided either by application experts or by offline estimation methods (e.g., \cite{NeuralKF_SS1,NeuralKF_SS2}). Besides potential for improved performance, these hybrid schemes offer improved interpretability over purely MF estimation mechanisms; a feature that is beneficial for safety critical monitoring applications. Additionally, NN-aided KFs with exact noise knowledge are discussed for example in references~\cite{NeuripsHybridInference,DANSE,KFwNeuralExtraction1}. For instance, the work in~\cite{NeuripsHybridInference} utilizes a two-step approach, where a conventional estimation algorithm is first applied, and then NNs perform error correction. A recent group of works focuses on the more generally applicable case of \textit{partial domain knowledge}, designing estimation filters without covariance information~\cite{kalmanNet,latentKalmanNet,stamatelis2025filteringjumpmarkovsystems}. In these works, a prior of the state is computed according to the KF, and then, specific features are extracted and used by RNNs to estimate the Kalman Gain (KG) without the availability of the noise covariance matrices. Then, that NN-based KG is used in the posterior filtering operation.    

In this paper, we focus on MA estimation systems and extend the aforedescribed NN-based, covariance-agnostic idea to distributed filtering. This distributed latent state estimation goal is actually more challenging due to the lack of global information and the induced communication costs. In particular, the agents/nodes performing estimation need to reach consensus (i.e., similar estimates) and not just minimize their local errors; to this end, consensus algorithms constitute an entire research sub-field with many intricacies~\cite{ConsensusGeneral1,ConsensusGeneral2}. In fact, even for linear distributed systems, optimal filters do not exist or are computationally prohibitive, forcing a reliance on heuristic consensus-based Kalman updates~\cite{KCF-saber}. Additionally, in the MA Reinforcement Learning (MARL) community, it is well known that MA optimization is more challenging due to the inherent nonstationarity of the learning process\textcolor{black}{\cite{CLDE_REF1,MADDPG}}. More specifically, as one agent updates its estimator model, it alters the distribution of the transmitted messages to neighboring nodes. Consequently, the environment dynamics of all other agents change, and previously learned behaviors must constantly adapt. In the following, we summarize the contributions of this paper:
\begin{itemize}
    \item A novel distributed estimator, termed as Covariance-Agnostic Neural Kalman Consensus Filter (CA-NKCF), is presented, which fuses principles from MARL with model-informed NN-based filtering and a novel consensus mechanism. A primary advantage of CA-NKCF is its lightweight communication structure, since, unlike advanced information-weighted filters\textcolor{black}{~\cite{KCF_types_CaseStudy,ICF}} that require exchanging full covariance or information matrices, the proposed approach requires only the exchange of state priors, substantially reducing the inter-agent/-node communication bandwidth. In addition, to overcome the inherent nonstationarity of MA optimization, the proposed NN filtering modules and consensus weights are jointly optimized offline using a central dataset. Inspired by the Centralized Learning, Decentralized Execution (CLDE) paradigm~\cite{CLDE_REF1,MADDPG}, the considered training allows the framework to discover effective collaborative behaviors, and promotes extreme scalability by sharing copies of  identical optimized NNs across nodes.
    \item We provide strong mathematical intuition regarding the stability of the proposed NN-based consensus mechanism. It is shown that, for each estimation node, the posterior estimate is a convex combination of their local prior and the priors of their neighboring nodes. This feature indicates that each local posterior is trapped inside the multi-dimensional convex hull of all priors, a fact that prevents catastrophic erroneous updates.
    \item We present extensive numerical experiments that showcase the superiority of the proposed CA-NKCF approach over a wide range of distributed estimators, including both MB and purely data-driven approaches, in both linear and nonlinear systems. Furthermore, a rigorous ablation study is used to demonstrate the critical importance of a unified loss function, a central training procedure, and the joint optimization of both the NN parameters and the consensus weights.
\end{itemize}
The remainder of this paper is organized as follows. A literature review of MB and  distributed filtering schemes is provided in Section~II, whereas the mathematical background of KFs and related algorithms is included in Section~III, along with the data-driven problem formulation under investigation. Our novel CA-NKCF approach is presented in Section~IV, and test performance is experimentally assessed in Section~V. Finally, Section VI includes the paper's concluding remarks.

\textbf{Notation:} Lower-case bold letters refer to vectors (e.g., $\mathbf{x}$), upper-case bold letters to matrices (e.g., $\mathbf{X}$), and calligraphic letters indicate sets (e.g., $\mathcal{X}$). $\mathcal{N}(\mu,\sigma)$ represents a Gaussian distribution with parameters mean $\mu$ and variance $\sigma$, whereas $\mathbb{E}[\cdot]$ denotes expectation. The notation $\mathbf{I}_d$ stands for the $d \times d$ ($d\geq2$) identity matrix, and $|\cdot|$ denotes the cardinality operator.

\section{Related Work}
\subsection{Conventional Distributed Filtering}
Decentralized implementations of the celebrated KF algorithms have been studied by the control community for over two decades, initially for ideal fully-connected networks~\cite{decKF_OLD_Rao}. The foundations for scalable decentralized KF with consensus algorithms for systems with sparse and possibly time-varying topologies have been set by Olfati-Saber in his seminal Kalman Consensus Filter (KCF) in~\cite{KCF-saber}. This algorithm combines local KF-like updates with a consensus imposed on the weighted sensor disagreement. Although this filter has not been proven to be optimal, due to the heuristic variance-dependent choice of the consensus weights, it remains very popular up to date due to its satisfactory performance, simplicity, and numerical stability.

Improvements that either run multiple consensus steps on the information vectors \cite{ICF} or compute the optimal value of consensus weights \cite{OKCF,OKF-IWC} have been proposed, at the expense, however, of much higher computational costs, practicality limitations, and stability risks. Computing optimal values as proposed in \cite{OKCF,OKF-IWC} requires maintaining and performing consensus on covariance matrices, and then performing multiple matrix inversions on massive block matrices for observation correlations, hindering numerical stability. 
All in all, even though these papers produce beautiful mathematical optimality results, Saber's KCF remains very popular due to its easy implementation and stability. 
Additionally,  progress has been made in combining distributed filters with practical engineering challenges, e.g., non-Gaussian observations \cite{DKF-nonGauss}, limited sensor range \cite{DKF_sensing_range}, communication costs \cite{DKF_comm_costs,DKF_Comm2}, as well as privacy risks \cite{DKF-Privacy}. 

It is noted that, while KF-based approaches are the most widely utilized estimators, an alternative family of methods known as Particle Filters (PFs) can be used in nonlinear systems instead. These filters constitute non-parametric sequential Monte Carlo estimators that have been used for single-agent~\cite{PF1,PF2,PFAI} as well as distributed, MA~\cite{DPF1} systems. However, these approaches carry much greater overhead and are highly sensitive to the number of particles simulated. 


\subsection{Model-Based Neural Filtering}
There are two main approaches for online hidden state estimation in discrete-time Dynamical Systems (DSs) using MB NNs: \textit{i}) external architectures; and \textit{ii}) NNs embedded in the KF logic. 
Prior works on \textit{i}), either utilize NNs to extract features from  high-dimensional observations, which are then combined with known state updates \cite{KFwNeuralExtraction1}, or utilize RNNs to perform error correction on traditional filters~\cite{NeuripsHybridInference}. Exact specification of the state evolution's mean and covariance is required. Most works on \textit{ii}) are based on the KalmanNet framework~\cite{kalmanNet}. This work first proposed utilizing supervised Gated Recurrent Units (GRUs)~\cite{GRU} to estimate the KG without the need for covariance matrix knowledge. The proposed hybrid model was shown to outperform both MB KFs and PFs as well as fully MF RNNs. Since then, various extensions of the original KalmanNet framework for different types of DSs have been proposed. For instance,~\cite{latentKalmanNet} combines the KalmanNet framework with deep convolutional feature extractors for tracking problems profiting from visual observations from cameras. Very recently, an extension of the KalmanNet with an additional GRU, termed as MJFNet~\cite{stamatelis2025filteringjumpmarkovsystems}, was designed to filter trajectories with switching behavior. One alternative framework in this line of research is the fully unsupervised Data-Driven Nonlinear State Estimation (DANSE) model~\cite{DANSE}, which does not require any specification of the latent state evolution model. However, it is limited to linear observations with fully known Gaussian noise. 

MB deep learning techniques for latent space estimation combine domain knowledge of traditional algorithms with the expressiveness and generalization capabilities of modern NNs to improve performance. 
MB deep learning has been also proposed for other tasks like smoothing, i.e., offline/noncausal state estimation~\cite{NeuripsHybridInference,kalmanNetSmoothing,smoothingAAAI}. Besides discriminative learning problems, NNs can also be used to estimate the underlying dynamics of high-dimensional nonlinear DSs, which is a form of generative learning~\cite{deepKFs,NeuralKF_SS1,NeuralKF_SS2,NeuralKF_SS3}. Note that discriminative and generative approaches can be used in parallel, i.e., by first learning the dynamics offline with a generative algorithm and then training a discriminative NN-based KF. In that case, the learned dynamics constitute the state and observation generating recursions utilized by the filter. It is, however, highlighted that all aforementioned works are \textit{limited to centralized single-node systems}, where a single processing center infers hidden state information using all available observations.

It is finally noted for completeness that, apart from estimation of time-varying processes, MB deep learning~\cite{MBDLSurvey} has numerous interesting applications, e.g., medical imaging~\cite{MBDL1}, near-field localization~\cite{MBDL2}, as well as dimensionality reduction~\cite{MBDL3}. An important subcategory of MB deep learning is physics-informed NNs, where physical laws are directly integrated into the NN operation, e.g., \cite{PhysInfDL,PhysInfDRL2,Graph_CNN,PhysInfDL3}.

\section{Preliminaries}
\subsection{Centralized Filters}
In the centralized setting (single node/sensor/agent), discrete-time DSs (indexed by time $t$) are described as follows:
\begin{subequations}\label{eq:DS_def_basic}
    \begin{align}
    \mathbf{x}_{t+1} &= f(\mathbf{x}_t)+\mathbf{w}_t \in \mathbb{R}^s\label{eq:DS_def_basic_a},\\
    \mathbf{z}_t&=h(\mathbf{x}_t)+\mathbf{v}_t \in \mathbb{R}^o.
\end{align}
\end{subequations}
where $f(\cdot)$ and $h(\cdot)$ are termed as the transition and observation functions, respectively, whereas $\mathbf{w}_t$ and $\mathbf{v}_t$ are noise vectors.
MB estimators typically assume that $\mathbf{w}_t\sim \mathcal{N}(0,\mathbf{Q}),\mathbf{v}_t \sim \mathcal{N}(0,\mathbf{R})$, where $\mathbf{Q} \in \mathbb{R}^{s\times s}, \mathbf{R} \in \mathbb{R}^{o \times o}$.

A simple yet fundamental category of DSs is the linear DS, where the functions $f(\cdot)$ and $h(\cdot)$ are respectively the matrices $\mathbf{F}$ and $\mathbf{H}$, yielding the state-space equations:
\begin{subequations}
    \begin{align}
    \mathbf{x}_{t+1} &= \mathbf{F} \mathbf{x}_t +\mathbf{w}_t \in \mathbb{R}^s,\\
\mathbf{z}_t&=\mathbf{H} \mathbf{x}_t+\mathbf{v}_t \in \mathbb{R}^o.
\end{align}
\end{subequations}
We are concerned with filtering, i.e., estimating the current latent variable $\mathbf{x}_t$, leveraging the past and present observations $\mathbf{z}_{1:t}\triangleq[\mathbf{z}_1,\mathbf{z}_2,\ldots, \mathbf{z}_t]$.  The prior and posterior estimates at the time instance $t$ are respectively denoted as follows: \begin{align}
    \hat{\mathbf{x}}_{t|t-1}\triangleq\mathbb{E}[\mathbf{x}_t|\mathbf{z}_{1:t-1}], 
    \, \hat{\mathbf{x}}_{t|t} \triangleq \mathbb{E}[\mathbf{x}_t|\mathbf{z}_{1:t}].
\end{align}
Consequently, the prior and posterior errors are denoted as $
\boldsymbol{\eta}_{t|t-1}$ and $\boldsymbol{\eta}_{t|t}
$, and the respective error covariance matrices are defined as $\mathbf{P}_t\triangleq\mathbb{E}[ \boldsymbol{\eta}_{t|t-1} \boldsymbol{\eta}_{t|t-1}^T  ]$ and $\mathbf{M}_t\triangleq\mathbb{E}[\boldsymbol{\eta}_{t|t} \boldsymbol{\eta}_{t|t}^T]$. The KF for linear DSs is given by the following recursion: 
\begin{subequations}\label{eq:KFRecursion}
    \begin{align}
    \mathbf{K}_t&=\mathbf{P}_t \mathbf{H}^T(\mathbf{R}+\mathbf{H} \mathbf{P}_t \mathbf{H}^T)^{-1}, \label{eq:singleKG}\\
    \hat{\mathbf{x}}_{t|t} &=\hat{\mathbf{x}}_{t|t-1}+\mathbf{K}_t (\mathbf{z}_t - \mathbf{H} \hat{\mathbf{x}}_{t|t-1}),\label{eq:posterior_update_singleKG} \\
    \mathbf{M}_t &= \mathbf{P}_t - \mathbf{P}_t \mathbf{H}^T (\mathbf{R}+\mathbf{H} \mathbf{P}_t \mathbf{H}^T)^{-1} \mathbf{H} \mathbf{P}_t,\\
    \mathbf{P}_{t+1} &=\mathbf{F} \mathbf{M}_t\mathbf{F}^T + \mathbf{Q},\\
    \hat{\mathbf{x}}_{t+1|t}&=\mathbf{F}\hat{\mathbf{x}}_{t|t}.
\end{align}
\end{subequations}
The variable $\mathbf{K}_t$ in expression~(\ref{eq:singleKG}) is known as the KG. The KalmanNet framework \cite{kalmanNet,latentKalmanNet,stamatelis2025filteringjumpmarkovsystems} performs the same a priori estimate for $\hat{\mathbf{x}}_{t|t-1}$ as the KF, but utilizes a GRU $\theta$ in order to estimate $\mathbf{K}_t$.  The approximate KG $\mathbf{K}_{t,\theta}$ is then plugged into the expression~(\ref{eq:posterior_update_singleKG}). It is noted for completeness that the EKF is a popular yet suboptimal approach for filtering nonlinear DSs, where the state updates become as follows:  
\begin{align}
&\hat{\mathbf{x}}_{t|t}=\hat{\mathbf{x}}_{t|t-1}+\mathbf{K}_t(\mathbf{z}_t-h(\hat{\mathbf{x}}_{t|t-1})),\,
\hat{\mathbf{x}}_{t+1|t}=f(\hat{\mathbf{x}}_{t|t}).
\end{align}

For the covariance updates and the KG computation, matrices $\mathbf{F}$ and $\mathbf{H}$ are replaced  by the following Jacobians which are evaluated on the state estimates:
\begin{align}
    \tilde{\mathbf{F}}_t=\mathbf{J}_f({\hat{\mathbf{x}}}_{t-1|t-1}), \, \tilde{\mathbf{H}}_t=\mathbf{J}_h({\hat{\mathbf{x}}}_{t|t-1}).
\end{align} 
It needs to be noted that the KalmanNet algorithm~\cite{kalmanNet} supports the use of EKF-like state predictions.

\subsection{The Kalman Consensus Filter (KCF)}
Consider an environment with $N$ sensors (indexed with $i=1,2,\ldots,N$) which collect distinct, possibly overlapping, measurements according to the following model:
\begin{equation}\label{eq:DS_def_basic_multi}
    \mathbf{z}_{i,t}=\mathbf{H}_i \mathbf{x}_t +\mathbf{v}_{i,t} \in \mathbb{R}^{o_i}, 
\end{equation}
where the covariance matrix of each $i$-th observation noise is represented as $\mathbf{R}_i \in \mathbb{R}^{o_i \times o_i}$. It is assumed that each sensor node $i$ can exchange messages with a set of neighbors $\mathcal{N}_{i,t}$; furthermore, we write $\mathcal{J}_{i,t}\triangleq\mathcal{N}_{i,t} \cup \{i\}$. The nodes are essentially located in a graph, where edges indicate sensor communication (i.e., message exchange). We assume that the graph topology can evolve with time, and is not controlled by the nodes. Algorithms that combine state estimation models with intelligent topology design are a very interesting potential research direction, outside the scope of this work.

The seminal work in~\cite{KCF-saber} proposed KCF according to which, each sensor node $i$ broadcasts its local prior state estimate, and the Kalman posterior update is combined with a consensus update rule based on the prior estimates. The update recursion of KCF is defined as follows:
\begin{subequations}
    \label{eq:KCF}\begin{align}
        &\mathbf{K}_{i,t}=\mathbf{P}_{i,t} \mathbf{H}_{i}^T (\mathbf{R}_i + \mathbf{H}_i \mathbf{P}_i \mathbf{H}_i^T)^{-1}, \label{eq:KDF_KG}\\
        & \hat{\mathbf{x}}_{i,t|t} = \hat{\mathbf{x}}_{i,t|t-1}+\mathbf{K}_{i,t} (\mathbf{z}_{i,t}-\mathbf{H}_i \hat{\mathbf{x}}_{i,t|t-1}) 
         \nonumber\\
        &\quad \quad \quad+ \mathbf{C}_{i,t} \sum_{j \in \mathcal{N}_i} (\hat{\mathbf{x}}_{j,t|t-1}-\hat{\mathbf{x}}_{i,t|t-1}), 
        \label{eq:posteriorConsensusKCF}\\
        &\mathbf{A}_{i,t}=\mathbf{I}-\mathbf{K}_{i,t} \mathbf{H}_i,\\
        &\mathbf{M}_{i,t}=\mathbf{A}_{i,t} \mathbf{P}_{i,t}\mathbf{A}_{i,t}^T+\mathbf{K}_{i,t} \mathbf{R}_i \mathbf{K}_{i,t},\\
        &\mathbf{P}_{i,t+1}=\mathbf{F}\mathbf{M}_{i,t}\mathbf{F}^T+\mathbf{Q},\\
        &\hat{\mathbf{x}}_{i,t+1|t}=\mathbf{F}\hat{\mathbf{x}}_{i,t|t},
    \end{align}
\end{subequations}
where a typical heuristic choice for the consensus weight is: \begin{equation}
    \label{eq:SaberHeuristicCi}
    \mathbf{C}_{i,t} \triangleq \epsilon \frac{\mathbf{P}_{i,t}}{1+||\mathbf{P}_{i,t}||_{\rm F}} \in \mathbb{R}^{s \times s}
\end{equation}
with $\epsilon$ being an appropriately chosen hyperparameter.

It is noted that, while the baseline recursion in~\eqref{eq:KCF} relies on a heuristic consensus matrix and the exchange of only prior state estimates, a rigorously derived variant in~\cite[Algorithm 3]{KCF_types_CaseStudy} offers improved performance. This advanced implementation employs an information-weighted formulation where nodes broadcast their local information vector ($\mathbf{u}_{i,t}$) and information matrix ($\mathbf{U}_{i,t}$) alongside their state predictions. By aggregating this data from the neighborhood $\mathcal{J}_{i,t}$, the distributed update successfully minimizes estimate disagreement across the MA network and bypasses the traditional KG matrix calculation. The complete recursion for this formulation is defined as:
\begin{subequations}
    \label{eq:KCF_Alg3}
    \begin{align}
        &\mathbf{u}_{j,t} = \mathbf{H}_j^T \mathbf{R}_j^{-1} \mathbf{z}_{j,t}\,\,\forall j \in \mathcal{J}_{i,t}, \\
        &\mathbf{U}_{j,t} = \mathbf{H}_j^T \mathbf{R}_j^{-1} \mathbf{H}_j \,\, \forall j \in \mathcal{J}_{i,t}, \\
        &\mathbf{y}_{i,t} = \sum_{j \in \mathcal{J}_{i,t}} \mathbf{u}_{j,t}, \\
        &\mathbf{S}_{i,t} = \sum_{j \in \mathcal{J}_{i,t}} \mathbf{U}_{j,t}, \\
        &\mathbf{M}_{i,t} = \left(\mathbf{P}_{i,t|t-1}^{-1} + \mathbf{S}_{i,t}\right)^{-1}, \\
        &\hat{\mathbf{x}}_{i,t|t} = \hat{\mathbf{x}}_{i,t|t-1} + \mathbf{M}_{i,t} \left( \mathbf{y}_{i,t} - \mathbf{S}_{i,t} \hat{\mathbf{x}}_{i,t|t-1} \right), \nonumber \\
        &\quad\quad\quad+ \epsilon \mathbf{M}_{i,t} \sum_{j \in \mathcal{N}_i} \left( \hat{\mathbf{x}}_{j,t|t-1} - \hat{\mathbf{x}}_{i,t|t-1} \right), \label{eq:posterior_alg3} \\
        &\mathbf{P}_{i,t+1|t} = \mathbf{F} \mathbf{M}_{i,t} \mathbf{F}^T + \mathbf{Q}, \\
        &\hat{\mathbf{x}}_{i,t+1|t} = \mathbf{F} \hat{\mathbf{x}}_{i,t|t}.
    \end{align}
\end{subequations}

\subsection{Estimation Problem Formulation}
In this work, we wish to design a data-driven distributed filtering algorithm, such that each sensor node $i$ can reliably estimate the true state $\mathbf{x}_t$, based on past observations and received messages. Formally, node $i$ employs an NN $\theta_i$ in order to estimate the latent state vector as: \begin{equation}
    \label{eq:NeuralConditionalEstimDef}
    \hat{\mathbf{x}}_{i,t|t}= \mathbb{E}[\mathbf{x}_t|\mathbf{z}_{i,1:t};\theta_i].
\end{equation}
In particular, the core objective is to minimize the average Mean Squared Error (MSE) loss; in mathematical terms: \begin{equation}\label{eq:corr_objective}
    \min_{\theta_1,\theta_2,\ldots,\theta_N}\frac{1}{N}\sum_{i=1}^N \mathbb{E}\left[\left\|\hat{\mathbf{x}}_{i,t|t}-\mathbf{x}_t\right\|^2\right].
\end{equation}

Following the reasoning of established works \cite{kalmanNet,kalmanNetSmoothing}, we proceed adopting the following realistic assumptions:
\begin{itemize}
    \item The noise distribution is unknown. In practical systems, noise distributions are often complex and nonstationary, and thus hard to estimate, necessitating filters that do not rely on the exact specification of noise statistics.    
    \item Approximations of the transition function $f(\cdot)$ in~\eqref{eq:DS_def_basic_a} and the observation functions $h_i(\cdot)$ $\forall i$ in the general version of~\eqref{eq:DS_def_basic_multi}: $\mathbf{z}_{i,t}=h_i(\mathbf{x}_t) +\mathbf{v}_{i,t}$ are available, either estimated offline or provided by an application expert. However, in the results section, we will investigate the effectiveness of our scheme when these approximations are incorrect. 
\end{itemize}
Furthermore, in order to train our NN models, we will assume the availability of a large labeled dataset with hidden states and sensor-wise observations, defining the following set: 
\begin{equation}
    \label{eq:DatasetDefinition}
    \mathcal{D}\triangleq\{ 
    \mathbf{x}_{1:t}^{(d)}, \mathbf{z}^{(d)}_{1,1:t},\mathbf{z}^{(d)}_{2,1:t},\ldots,\mathbf{z}_{N ,1:t}^{(d)}
    \}_{d=1}^{D}.
\end{equation}
Finally, the design of our distributed estimator, described in the sequel, is influenced by two critical practical requirements:

\begin{itemize}
    \item[R1] \textbf{Strict Real-Time Causality:} In highly dynamic systems, the state transitions from $\mathbf{x}_t$ to $\mathbf{x}_{t+1}$ vary rapidly. To prevent predictions from becoming obsolete, the estimator's forward inference time needs to be very quick. This hard latency constraint precludes the use of computationally heavy architectures (e.g., massive transformers~\cite{attention} or graph NNs~\cite{Graph_CNN}), high-capacity PFs~\cite{DPF1}, or iterative multi-step consensus protocols.
    \item[R2] \textbf{Massive Scalability:} The designed framework needs to be readily deployable in large-scale Internet of Things (IoT) applications comprising a vast number of sensor nodes. This necessitates an architecture that inherently minimizes memory footprint and training complexity, a requirement that directly motivates the Parameter Sharing (PS) strategy introduced in the sequel.
\end{itemize}

\begin{remark}[Topology Agnosticism and Overfitting]
A critical challenge in data-driven distributed estimation is the risk of the machine learning model overfitting to a specific, favorable message-exchange graph. To this end, in this paper, we deliberately do not assume any fixed topology structure. The proposed framework is required to be inherently topology-agnostic, ensuring that the nodes learn generalized collaborative filtering behaviors rather than indirectly inferring and exploiting static connection patterns. To rigorously validate this requirements, the ensuing numerical experiments in Section~\ref{sec:Results} utilize random graphs drawn from a maximum entropy distribution (i.e., uniform connection probabilities at each time instance $t$).
\end{remark}

\section{Proposed Distributed Estimation Method}
Having established the necessary mathematical background and presented the hybrid machine learning setup, we will now present our proposed CA-NKCF.
We have adopted a PS method where all $N$ sensor nodes are equipped with the same NN parameters $\theta$ (i.e., $\theta_i=\theta$ $\forall i$); this choice was actually made for scalability purposes. In modern sensor network applications, the number of sensors can be very large, implying that initializing and optimizing $N$ separate RNNs carries prohibitive computational and memory costs. By adopting PS in our multi-node setup, only one RNN needs to be trained, which is very cost efficient and memory friendly. Notably, PS is often used for large-scale MARL \cite{DotaDRL,PSPruning,PSRLRobotics} for the same reason. Furthermore, in distributed estimation, in addition to total error minimization, the nodes must reach similar state estimates. Since the NN parameters  are shared,  unless the local inputs differ significantly, we expect the disagreement to be minimal. In the next section, we will  demonstrate that CA-NKCF achieves better consensus than MB baselines.
\begin{algorithm}[!t]
    \caption{Proposed CA-NKCF at each $i$-th Node}
    \label{alg:ca_dnkf}
    \begin{algorithmic}[1] 
     \item[]\hspace{-0.3cm}\textbf{Input}: Observation $\mathbf{z}_{i,t}$, shared parameters $\theta$ and $\bm{\gamma}$, neighbors $\mathcal{N}_{i,t}$, and previous GRU hidden state $\mathbf{h}_{i,t-1}$. 
    \item[] \hspace{-0.3cm}\textbf{Output}: Posterior $\hat{\mathbf{x}}_{i,t|t}$ and new hidden state $\mathbf{h}_{i,t}$.
        \item[] \textbf{Local Prediction:}
        \STATE Compute the local prior $\hat{\mathbf{x}}_{i,t|t-1} = f(\hat{\mathbf{x}}_{i,t-1|t-1})$.
        \STATE Broadcast $\hat{\mathbf{x}}_{i,t|t-1}$ to neighbors $j \in \mathcal{N}_{i,t}$.
        \item[]\textbf{Neural KG Estimation:}
        \STATE Compute the RNN features as in expression~\eqref{eq:features}.
        \STATE Set $\mathbf{K}_{i,t;\theta},  \mathbf{h}_{i,t} = \mathsf{FP}_{\theta}(\boldsymbol{\phi}_{i,t}; \mathbf{h}_{i,t-1})$.
        \item[] \textbf{Consensus-Based Update:}
        \STATE Collect all prior estimates $\hat{\mathbf{x}}_{j,t|t-1}$ $\forall j \in \mathcal{N}_{i,t}$.
        \STATE Compute the consensus term:\\ 
         \hspace{1cm}$\mathbf{u}_{i,t}^{\text{cons}} = \frac{1}{|\mathcal{N}_{i,t}|}\sum_{j \in \mathcal{N}_{i,t}} \left(\hat{\mathbf{x}}_{j,t|t-1} - \hat{\mathbf{x}}_{i,t|t-1}\right)$.
        \STATE Perform the update: \\
        \hspace{1cm}$\hat{\mathbf{x}}_{i,t|t} = \hat{\mathbf{x}}_{i,t|t-1} + \mathbf{K}_{i,t;\theta} \Delta \hat{\mathbf{z}}_{i,t} + \sigma(\bm{\gamma}) \mathbf{u}_{i,t}^{\text{cons}}.$
        \STATE \textbf{return} $\hat{\mathbf{x}}_{i,t|t}$ and $\mathbf{h}_{i,t}$.
    \end{algorithmic}
\end{algorithm}

To remain consistent with prior works, we model $\theta$ as GRU, but other models, such as Long Short Term Memory (LSTM) models \cite{LSTM} or transformers~\cite{attention}, can be used instead. At each time instance $t$ and for each sensor node $i$, the NN parametrized by $\theta$ is provided with KF-specific input features $\boldsymbol{\phi}_{i,t}$, and then combines them with its most recent internal hidden state $\mathbf{h}_{i,t-1}$ to estimate the local KG $\mathbf{K}_{i,t;\theta}$.
\begin{figure*}
    \centering
    \includegraphics[width=\linewidth]{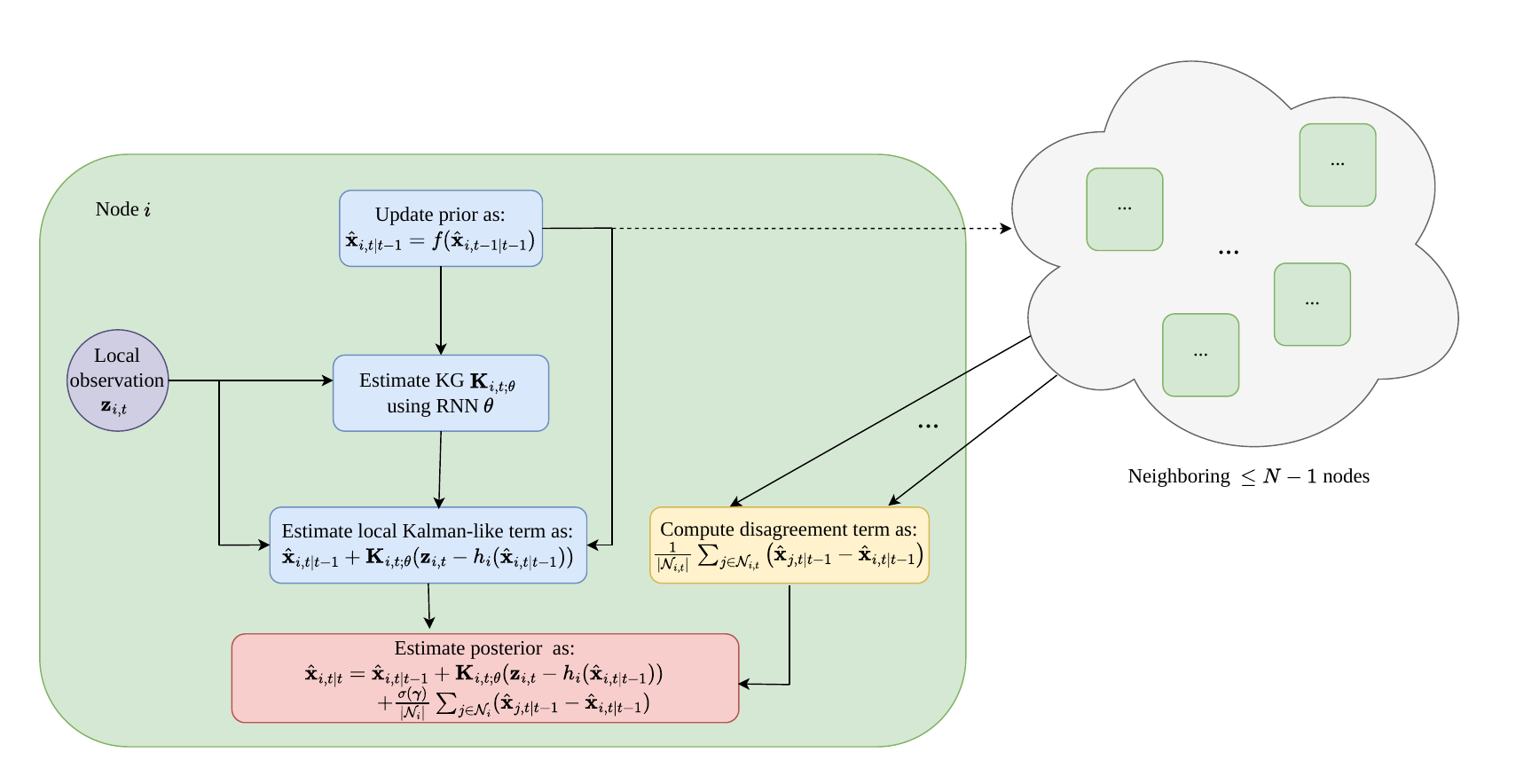}
    \caption{Visualization of the proposed CA-NKCF framework for distributed estimation with $N$ sensor nodes.}
    \label{fig:viz}
\end{figure*}
\subsection{Covariance-Agnostic Neural Kalman Consensus Filtering}
At each time instance $t$, each sensor node $i$ performs the following steps, collectively presented in Algorithm~\ref{alg:ca_dnkf} and visualized in Fig.~\ref{fig:viz}.  First, the prior state estimate $\hat{\mathbf{x}}_{i,t|t-1}=f(\hat{\mathbf{x}}_{i,t-1|t-1})$ is computed and transmitted to its neighboring nodes.
Then, the RNN input features are computed as follows:
\begin{equation}
    \label{eq:features}
        \boldsymbol{\phi}_{i,t} \triangleq \left[\Delta \hat{\mathbf{z}}_{i,t},\Delta \bar{\mathbf{z}}_{i,t},\Delta \hat{\mathbf{x}}_{i,t}\right] \in \mathbb{R}^{2o_i+s \times 1},
\end{equation}
including the following vectors:
\begin{subequations}
    \begin{align}
        \Delta \hat{\mathbf{z}}_{i,t}& \triangleq \mathbf{z}_{i,t}-h_i(\hat{\mathbf{x}}_{i,t|t-1}) \in \mathbb{R}^{o_i \times 1},\\
        \Delta \bar{\mathbf{z}}_{i,t}& \triangleq \mathbf{z}_{i,t}-\mathbf{z}_{i,t-1}
        \in \mathbb{R}^{o_i \times 1},
        \\
        \Delta \hat{\mathbf{x}}_{i,t}& \triangleq \hat{\mathbf{x}}_{i,t-1|t-1}-\hat{\mathbf{x}}_{i,t-1|t-2} \in \mathbb{R}^{s \times 1}. 
    \end{align}
\end{subequations}
This feature combination is selected due to its demonstrated effectiveness in prior single-sensor works \cite{kalmanNet,stamatelis2025filteringjumpmarkovsystems}. The KG is estimated using the GRU as $\mathbf{K}_{i,t;\theta}=\mathsf{FP}_{\theta} (\boldsymbol{\phi}_{i,t}; \mathbf{h}_{i,t-1})$, where $\mathbf{h}_{i,t-1}$ is the previous hidden state of the GRU and $\mathsf{FP}_v(\cdot;\mathbf{h})$ denotes the forward-pass operation of an RNN with parameters $v$ and hidden state $\mathbf{h}$. For each node $i$, a separate hidden state $\mathbf{h}_{i,t}$ is maintained and updated over time. Using a consensus weight $\gamma_{\xi}$ for each $\xi$-th possible state (with  $\xi \in \{1,2,\cdots,s\}$) included in the consensus vector $\boldsymbol{\gamma}$, the posterior estimate is:
\begin{align}
        \hat{\mathbf{x}}_{i,t|t} =& \hat{\mathbf{x}}_{i,t|t-1}+\mathbf{K}_{i,t;\theta} (\mathbf{z}_{i,t}-h_i (\hat{\mathbf{x}}_{i,t|t-1})) 
         \nonumber\\
        &+ \frac{\sigma(\boldsymbol{\gamma})}{|\mathcal{N}_i |} \sum_{j \in \mathcal{N}_i} (\hat{\mathbf{x}}_{j,t|t-1}-\hat{\mathbf{x}}_{i,t|t-1}) 
        \label{eq:posteriorConsensusKCF},
    \end{align}
where $\sigma(\cdot)$ is an element-wise sigmoid function having the role of forcing the consensus weight inside $(0,1)$ to prevent numerical issues. Note that, in this expression, instead of utilizing a large consensus matrix $\mathbf{C}_{i,t}\in \mathbb{R}^{s \times s}$ as in~\cite{KCF-saber}, we employ the single learnable consensus weight $\gamma_\xi$ for each $\xi$-th state, which models each node's trust on its own information versus the information provided by the peers regarding that specific state component. We argue that the complexity of handling nonlinearities, correlations, and elaborate dynamics is sufficiently managed by the proposed neural KG estimation and, hence, decoupled dimension-wise consensus is entirely sufficient; this will become evident in the numerical investigations presented later on in Section~\ref{sec:Results}. Consequently, the optimization process only needs to learn the general informative value and reliability of the shared peer estimates for each individual state variable.

\begin{remark}[State-of-the-art vs. Proposed Consensus Scheme]
Arguably, optimizing a full $s \times s$ consensus matrix as presented in~\cite{KCF-saber} drastically expands the parameter search space, a fact that introduces a significant risk of learning spurious cross-dimensional couplings. Consider, for example, a $2$-Dimensional (2D) vehicle tracking application where the state vector comprises Cartesian positions ($x, y$) alongside their respective velocities ($v_x, v_y$).
From a physics standpoint, a velocity on one axis does not affect the position of another axis, and the model should not impose such correlations. 
Because a fully populated matrix lacks inherent kinematic constraints, off-diagonal elements can easily learn to cross-contaminate orthogonal axes. For instance, a sudden lateral position displacement reported by a neighbor could be erroneously mapped into a longitudinal acceleration correction for the local sensor node. Since these cross-terms violate the true decoupled equations of motion, they frequently lead to instability and filter divergence. By restricting the consensus interactions to a decoupled learnable vector $\boldsymbol{\gamma}$ within the proposed CA-NKCF framework (see \eqref{eq:posteriorConsensusKCF}), we enforce dimension-wise independence and eliminate this physically inconsistent cross-contamination risk.
\end{remark}

\begin{remark}[Computational and Communication Efficiency]
The proposed filtering approach improves the numerical stability and communication costs of the conventional KCF~\cite{KCF-saber}. First, traditional KG computation  requires matrix inversion, which carries cubic cost and is associated with increased risks of numerical instability. This issue is exacerbated  when Jacobian-based matrices are used for nonlinear systems, and additional stability mechanisms need to be employed further increasing the cost.
In contrast, in the proposed CA-NKCF framework, each sensor node $i$ only performs a prior estimate using $f(\cdot)$ (or $\mathbf{F}$ in the linear case) and a forward pass of a moderate-sized GRU.
Besides that, in its more advanced form, the KCF requires exchange of  two $s$-dimensional vectors and an $s\times s$ matrix, whereas our algorithm only requires messaging an $s$-dimensional prior. Thereby, it allows for implementation with substantially lower bandwidth consumption.
\end{remark}

\subsection{Mathematical Intuition Regarding Stability}
Providing a rigorous stability proof for a distributed recurrent estimator is a difficult task that falls beyond the scope of this paper. However, we herein provide strong mathematical intuition as to why we expect the consensus mechanism within the proposed CA-NKCF framework to be stable in practice. Focusing on the proposed update rule in~\eqref{eq:posteriorConsensusKCF} for a single dimension $\xi$, the following is deduced: 
\begin{align}
     &\hat{\mathbf{x}}_{i,t|t}[\xi] = \hat{\mathbf{x}}_{i,t|t-1}[\xi] + \frac{\sigma(\gamma_\xi)}{|\mathcal{N}_i|} \sum_{j \in \mathcal{N}_i} \left( \hat{\mathbf{x}}_{j,t|t-1}[\xi] - \hat{\mathbf{x}}_{i,t|t-1}[\xi] \right) \nonumber \\
    &= \underbrace{(1 -  \sigma(\gamma_\xi))}_{\text{self-weight}} \hat{\mathbf{x}}_{i,t|t-1}[\xi] + \sum_{j \in \mathcal{N}_i} \underbrace{\frac{\sigma(\gamma_\xi)}{|\mathcal{N}_i|}}_{\text{neighbor-weight}} \hat{\mathbf{x}}_{j,t|t-1}[\xi].\label{recursion-single} 
\end{align}
This decomposition implies the following significant properties of the proposed latent state estimation recursion: 
\begin{enumerate}
    \item \textbf{Non-negativity:} The sigmoid activation in~\eqref{recursion-single} guarantees that $\sigma(\gamma_\xi) \in (0,1)$, hence, the local self-weight and the individual neighbor weights incorporated into the update rule are strictly positive. Strict positivity is a well-established requirement for the stability of dynamic consensus systems \cite{RenAndBeard}.
    \item \textbf{Convexity:} It can be easily concluded that the latter weights sum to unity, implying that the posterior estimate is a convex combination of the local priors. Consequently, the state estimates are perpetually bounded within the multi-dimensional convex hull of the network's current ``beliefs''~\cite{SaberConsTutorial}, precluding compounding over-corrections and catastrophic divergence often observed in unconstrained end-to-end MA models. 
\end{enumerate}

It is finally noted that, beyond the proposed consensus step, we expect the local Kalman-like tracking to remain stable, as recurrent KG estimation using GRUs has been thoroughly studied and validated in the single-agent domain. Hence, while we cannot rigorously prove global stability and convergence for MA systems, due to the inherent complexity of GRUs, the synthesis of bounded consensus and reliable local filtering provides strong theoretical foundation for it. This intuition is further corroborated by the numerical experiments provided later on in the respective section.
\color{black}

\subsection{Training Optimization}
The loss function over the dataset $\mathcal{D}$ in~\eqref{eq:DatasetDefinition} corresponding to each $i$-th sensor node was chosen as follows:
\begin{equation}
    \mathcal{L}_i \triangleq \sum_{d=1}^D \sum_{t=1}^T \left\|\hat{\mathbf{x}}^{(d)}_{t|t-1}-\mathbf{x}_t^{(d)}\right\|^2,
\end{equation}
and the total, centralized loss function was set as the average of the local estimation errors. In this paper, we treat the learnable weights of NN $\theta$ and the consensus weights $\bm{\gamma}$ as a unified learnable parameter set, formulating the following optimization objective for the considered NN-based distributed estimation framework: 
\begin{equation}
    \label{eq:central_loss}
    \min_{\theta,\boldsymbol{\gamma}} \mathcal{L} \triangleq\frac{1}{N}\sum_{i=1}^N\mathcal{L}_i.
\end{equation}

The proposed model was trained on the mini-batch version of $\mathcal{L}$. For each trajectory in the batch, the neural filtering operation from $t=0$ up to $t=T$ was conducted sequentially, 
and the local losses at each time instance $t$ were added to compute the final loss result. For moderate trajectories (e.g., up to $T=100$), the hidden state vectors $\mathbf{h}_{1,t},\mathbf{h}_{2,t},\ldots,\mathbf{h}_{N,t}$ were initialized at the beginning, and the gradients were computed using Back Propagation Through Time (BPTT). For longer horizons, the trajectories can be split into shorter segments of length $T_{\rm trunc}$ to apply truncated BPTT. To handle datasets with trajectories of different lengths, max-padding can be used. The considered training process on a single trajectory is summarized in Algorithm~\ref{alg:loss_computation}.
\begin{algorithm}[!t]
    \caption{CA-NKCF Optimization on a Single Trajectory}
    \label{alg:loss_computation}
    \begin{algorithmic}[1]
    \item[] \hspace{-0.3cm}\textbf{Input}: Trajectory data $\mathbf{x}_{1:T}, \mathbf{z}_{1,1:T},\mathbf{z}_{2,1:t},\ldots,\mathbf{z}_{N,1:T}$ and neighbors $\{\mathcal{N}_{i,t}\}$. \\
    \item[] \hspace{-0.3cm}\textbf{Parameter}: Shared  parameters $\theta$ and $\bm{\gamma}$. \\
    \item[] \hspace{-0.3cm}\textbf{Output}: Optimized $\theta$ and $\bm{\gamma}$.
        \item[] \textbf{Initialization}
        \STATE Initialize GRU hidden states as $\mathbf{h}_{i,0} = \mathbf{0}$ $\forall i$.
        \STATE Initialize state estimates $\hat{\mathbf{x}}_{i,0|0}$. 
        \STATE Set $\mathcal{L}_{\text{traj}} = 0$.

        \FOR{$t = 1$ \textbf{to} $T$}
            \item[] \textbf{Distributed Filtering Step}
            \FOR{$i = 1$ \textbf{to} $N$}
                \STATE Call CA-NKCF (Algorithm \ref{alg:ca_dnkf}):\\
                 $\hat{\mathbf{x}}_{i,t|t}, \mathbf{h}_{i,t} \leftarrow \text{CA-NKCF}(\mathbf{z}_{i,t}, \theta, \bm{\gamma}, \mathcal{N}_{i,t}, \mathbf{h}_{i,t-1}).$
            \ENDFOR
            
            \item[] \textbf{Loss Accumulation}
            \STATE Compute the instantaneous loss:\\
             $\mathcal{L}_{ t} = \frac{1}{N} \sum_{i=1}^N \|\hat{\mathbf{x}}_{i,t|t} - \mathbf{x}_t\|^2.$ 
            \STATE Perform the update $\mathcal{L}_{\text{traj}} = \mathcal{L}_{\text{traj}} + \mathcal{L}_{ t}.$
        \ENDFOR
        \item[] \textbf{Parameter Optimization}
        \STATE Compute $\nabla_{\theta,\bm{\gamma}}{\mathcal{L}}_{\text{traj}}$ using BPTT.
        \STATE Update $\theta$ and $\bm{\gamma}$ using Adam \cite{Adam}.
        \STATE \textbf{return} $\theta,\bm{\gamma}$
    \end{algorithmic}
\end{algorithm}

\section{Numerical Results and Discussion}\label{sec:Results}
\begin{figure}[t]
    \centering
    {\includegraphics[width=\linewidth]{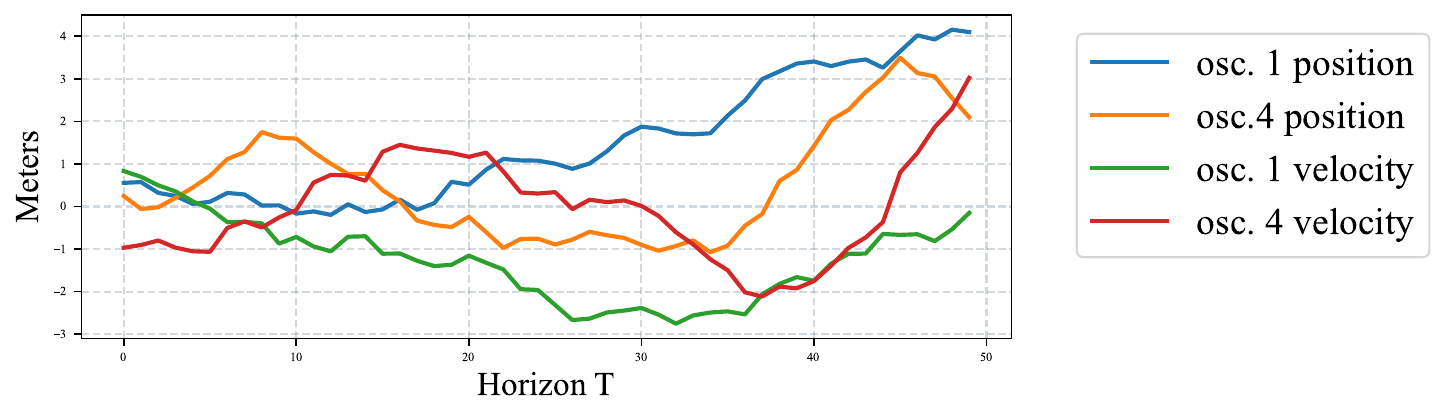}}
    \caption{Sampled oscillator states for $N=4$ sensors. }
    \label{fig:OscTraj}
\end{figure}
\begin{figure*}[t]
    \centering
    \begin{subfigure}[b]{0.32\textwidth}
        \centering
        \includegraphics[width=\linewidth]{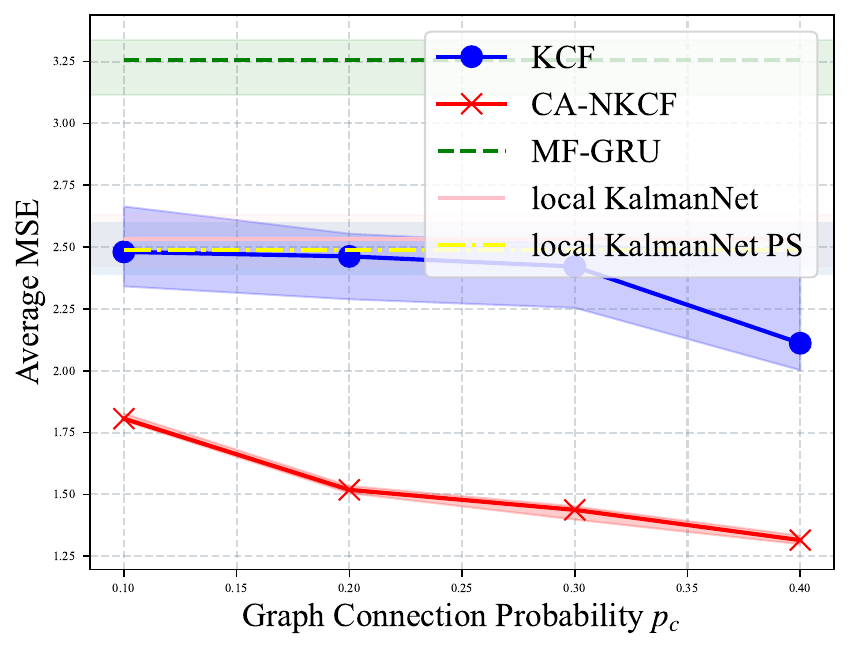} 
        \caption{Average Sensor MSE}
        \label{fig:avg_mse}
    \end{subfigure}
    \hfill 
    \begin{subfigure}[b]{0.32\textwidth}
        \centering
        \includegraphics[width=\linewidth]{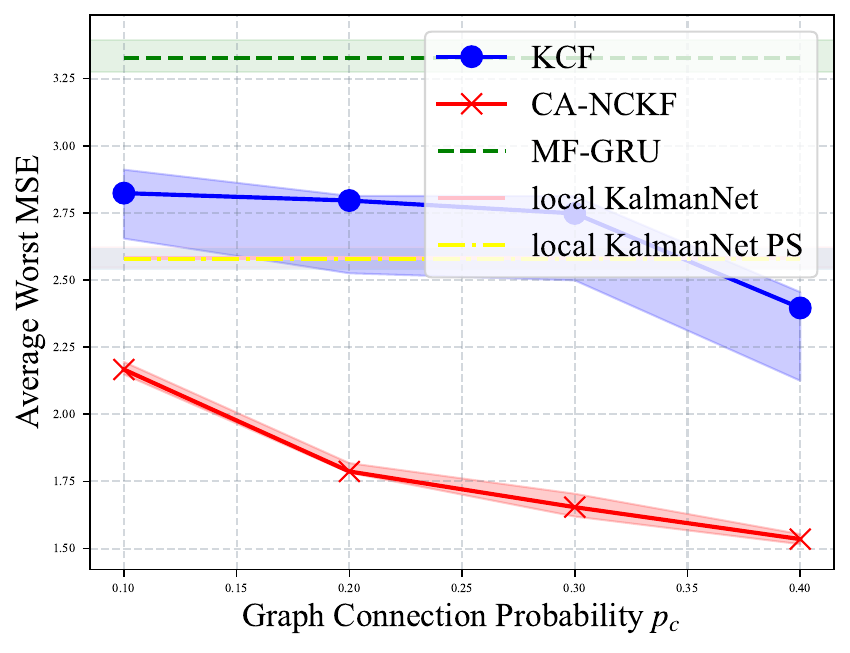} 
        \caption{Worst Sensor MSE}
        \label{fig:worst_mse}
    \end{subfigure}
    \hfill
    \begin{subfigure}[b]{0.32\textwidth}
        \centering
        \includegraphics[width=\linewidth]{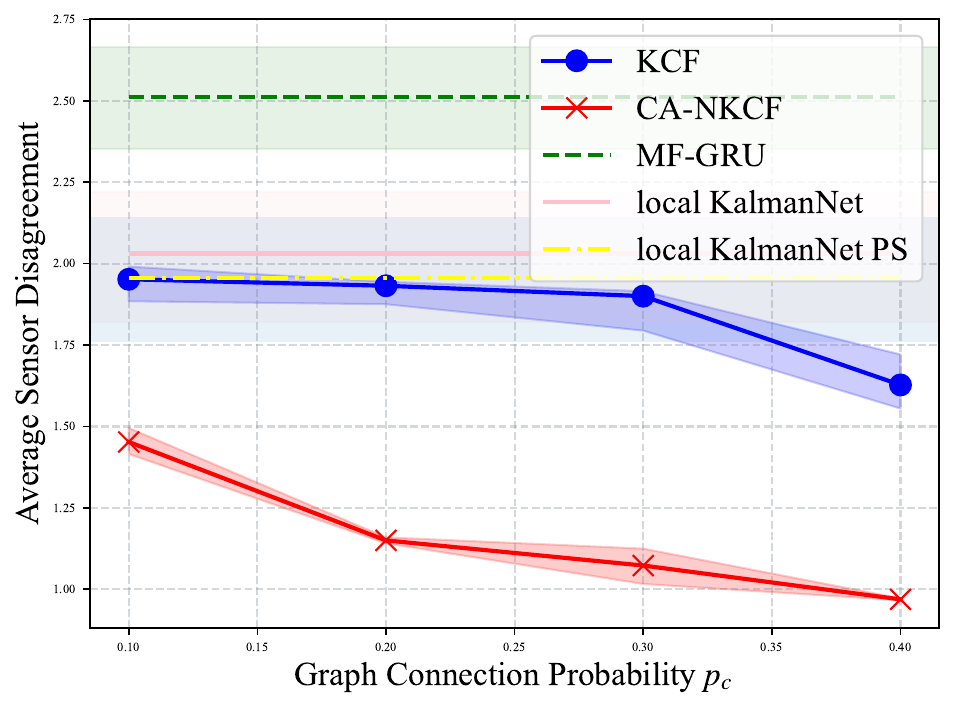} 
        \caption{Consensus Disagreement}
        \label{fig:disagreement}
    \end{subfigure}
    \caption{MSE performance versus the node connection probability for the linear scenario.}
    \label{fig:linearResultSmall}
\end{figure*}
In this section, diverse numerical investigations are presented to validate the effectiveness of the proposed CA-NKCF framework, by comparing it with traditional MB filters and MF NNs. We have simulated three scenarios: a linear example with harmonic oscillators, a nonlinear experiment inspired by chaos theory including a Lorenz attractor, and a practical wireless tracking application. For all simulations, the node/sensor/agent graph was chosen at random and time-varying, with the goal to verify that our model does not overfit to specific favorable topologies. For each time instance $t$, the existence of a communication link between nodes $i$ and $j$ was determined by an independent Bernoulli trial with probability $p_c$ (thus, determining\footnote{As emphasized in Remark 1, this dynamic, random topology generation is considered to showcase that the performance gains of the proposed CA-NKCF framework stem entirely from its robust consensus mechanism, precluding the GRU from profiting itself by memorizing static communication patterns that may not be present in real-world deployments.
} the sets of neighboring nodes $\mathcal{N}_{i,t}$ $\forall i$). 

All reported results in the sequel were obtained through averaging across $10$ random seeds; each instance was trained on $5 \times 10^4$ trajectories and tested on other $10^4$ ones. Training consisted of $100$ epochs performed with a learning rate of $5 \times 10^{-4}$. For the linear example, the CA-NKCF was realized with a GRU with $2$ hidden layers of $64$ units. For the nonlinear experiment, the hidden size was increased to $256$, and an additional pre-processing feed-forward module with hidden dimension of $128$ and a ReLU activation was used before the GRU. The output of the GRU was transformed to the appropriate KG dimensions with an additional linear layer. For MF benchmark, GRUs with similar structure trained on the same dataset were considered. Since they did not have access to domain knowledge, the  hidden sizes, and the number of hidden layers were doubled. For MB benchmarks, the KCF given by~(\ref{eq:KCF_Alg3}) was utilized for linear systems, and the Extended KCF (EKCF), Unscented KCF (UKCF) and Distributed PF (DPF) with $200$ particles were examined for nonlinear systems.

\subsection{Linear Scenario}
An environment with $N$ decoupled one-dimensional harmonic oscillators (same as the number of nodes) has been simulated. In particular, the state $\mathbf{x}_t \in \mathbb{R}^{2N}$ was a vector stacking the position and velocity of all oscillators. Each node $i$ observed only one oscillator, meaning that estimation of the entire state $\mathbf{x}_t \in \mathbb{R}^{2N}$ relied on successful consensus, and not just on developing powerful local KG estimators. The state evolution was actually dictated by the block-diagonal matrix $\mathbf{F}\triangleq\text{diag}(\mathbf{F}_1,\mathbf{F}_2,\ldots,\mathbf{F}_N)$, where each sub-block monitored a rotation with a distinct frequency, i.e., $\forall i$: \begin{equation}
    \mathbf{F}_i = \begin{bmatrix}
        \cos({\omega_i \Delta_t}) & -\sin(\omega_i \Delta_t) \\
        \sin(\omega_i \Delta_t) & \cos({\omega_i \Delta_t}) 
    \end{bmatrix}.
\end{equation} 

We have examined systems with $N=4k, k \in \mathbb{N} $ oscillators, setting $\Delta_t=0.1$ sec. The first quarter of the oscillators had the frequency $0.5 $ rad/sec, the second quarter $1$ rad/sec, the third $1.5$ rad/sec, and the final $2$ rad/sec. The process noise in~\eqref{eq:DS_def_basic} was chosen as zero-mean Gaussian with $\mathbf{Q}=0.05 \mathbf{I}_{2N}$. Each measurement matrix $\mathbf{H}_i \in \mathbb{R}^{2 \times 2N}$ extracted only the $i$-th component of the complete state and all observation noise matrices were set as $\mathbf{R}_i =0.1 \mathbf{I}_{2}$. Finally, he horizon was fixed to $T=50$. Examples of state evolution for the first and last oscillator for $N=4$ are depicted in Fig.~\ref{fig:OscTraj}.

\begin{figure}[!t]
    \centering
    \includegraphics[width=0.9\linewidth]{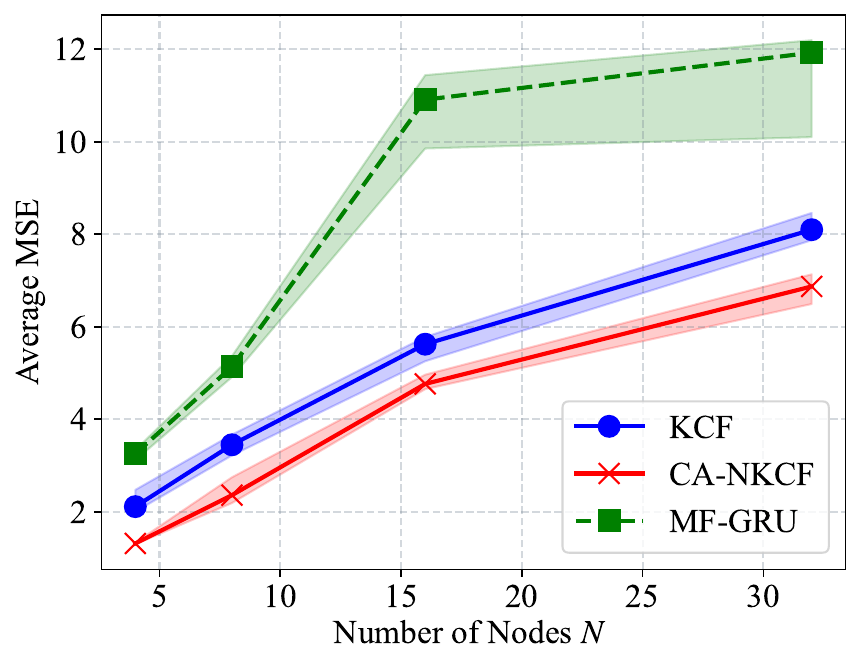}
    \caption{Average MSE versus the number of nodes $N$ for the linear scenario.}\vspace{-0.5cm}
    \label{fig:linearLarge}
\end{figure}

First, we considered a small system with $N=4$ nodes, varying $p_c$ from $0.1$ to $0.4$. As is evident from Fig.~\ref{fig:avg_mse}, the proposed method significantly outperforms all benchmarks. For further intuition, we have trained local KalmanNets for each node, both with shared (i.e., PS) and individual parameters. It is shown that these methods perform worse than the linear KCF, validating the paramount importance of consensus for this partially observable task. Interestingly, it is showcased that utilizing powerful NNs to perform KG estimation is not enough for distributed systems, indicating that the nodes must also perform intelligent consensus.
We also tracked the worst case MSE (out of all nodes) for each episode as well as the average disagreement in Figs.~\ref{fig:worst_mse} and~\ref{fig:disagreement} respectively, which collectively verify that our approach can successfully optimize these robustness metrics without being directly trained to do so. Finally, in Fig.~\ref{fig:linearLarge}, we fixed $p_c=0.4$ and varied $N$ up to $32$ nodes. As is apparent from the presented results, the proposed CA-NKCF scales well with $N$ consistently outperforming the baselines. Besides superior performance, our filter also achieves very good robustness to initialization conditions as inferred by examining the shaded region.  In contrast, the shaded region of the GRU in Fig. \ref{fig:linearLarge} is significantly larger.
\begin{table}[!t]
    \centering
    \begin{tabular}{lcccc}
        \toprule
        \textbf{Method} & $p_c=0.1$ & $p_c=0.2$ & $p_c=0.3$ & $p_c=0.4$ \\
        \midrule
        \textbf{ CA-NKCF} & \textbf{1.81} & \textbf{1.52} & \textbf{1.44} & \textbf{1.32} \\
        Ablation 1    & 2.01 & 1.67 & 1.57 & 1.56 \\
        Ablation 2  & 2.27 & 1.99 & 1.54 & 1.49 \\
        Ablation 3      & 2.11 & 1.88 & 1.71 & 1.62 \\
        \bottomrule
    \end{tabular}
     \caption{Ablation analysis for the proposed CA-NKCF in the linear scenario.}\label{tab:ablation_results}
\end{table}
\subsection*{Ablation Analysis}
We have verified the importance of each component of the prososed CA-NKCF by examining the following three ablation benchmarks: \begin{enumerate}
    \item To examine whether the KG computation is a necessary step, we have trained a GRU that processes the most recent observation, the prior $\hat{\mathbf{x}}_{i,t|t-1}$, and the average prior of all peers in the set $\mathcal{N}_{i,t}$, with the goal to directly map them to $\hat{\mathbf{x}}_{i,t|t}$. 
    \item  One important feature of our method is the joint optimization of $\bm{\gamma}$. To investigate whether joint optimization hinders stability, we have trained the same models, but with a fixed $\bm{\gamma}$, whose value was determined by an element-wise grid search on $[0,1]$. The training process was repeated for each considered value of $\bm{\gamma}.$
    \item For the final ablation test, we have investigated the importance of  end-to-end training with the consensus loop present during training, and the centralized loss function. To this end, we have trained $N$ independent local KalmanNets (one for each node), and then during testing alone, we added the consensus term in the recursion. Again, the value of the weight was determined by grid search on $[0,1]$. 
\end{enumerate}
For fairness, all latter $3$ GRUs were designed to have similar structure to the GRU of CA-NKCF. As depicted in Table~\ref{tab:ablation_results}, all $3$ components are necessary. The Kalman-like recursion is superior to just processing priors, optimizing $\bm{\gamma}$ along with $\theta$ improves training, and  the network must learn to cooperate during training by incorporating the consensus step in the training code; consensus on deployment alone is not sufficient.

\begin{figure}[!t]
    \centering
   \scalebox{0.9}{\includegraphics[width=\linewidth]{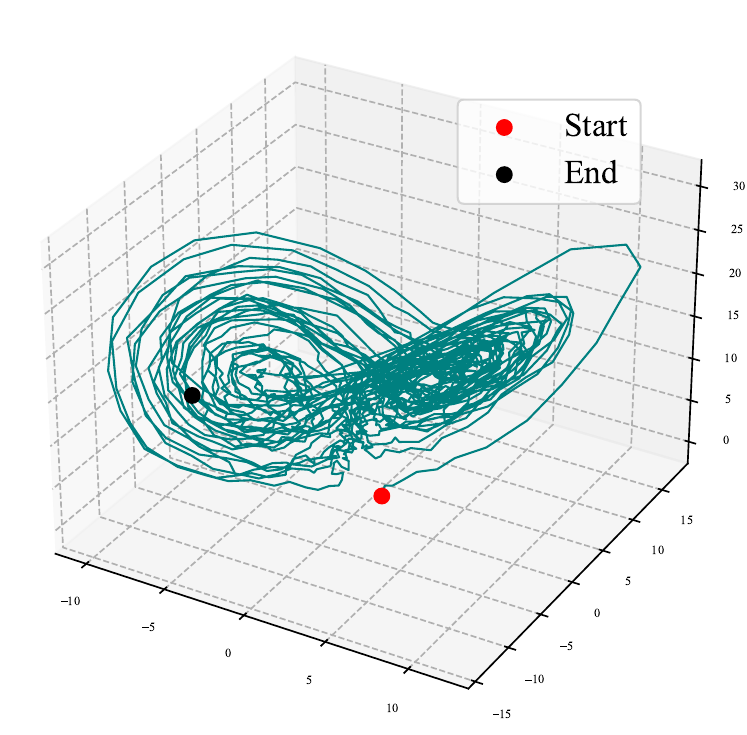}}
    \caption{The considered state trajectory for the Lorenz attractor scenario.}
    \label{fig:lorenzExample}
\end{figure}
\begin{figure}[!t]
    \centering
    \scalebox{0.9}{\includegraphics[width=\linewidth]{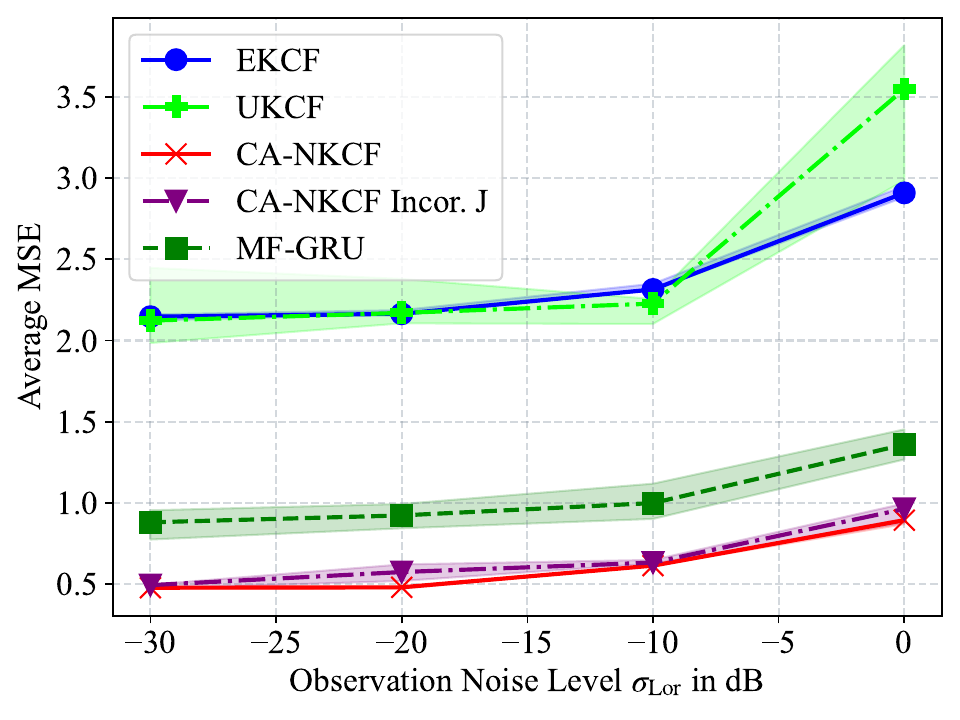}}
    \caption{Average MSE versus the observation noise level for the Lorenz attractor scenario.}
    \label{fig:lorenzResults}
\end{figure}
\subsection{Lorenz Attractor}
Next, we performed experiments with a polynomial $3$-dimensional DS inspired by the Lorenz attractor. Such DSs are often utilized to test the effectiveness of sequential estimators due to their inherent complexity. The latent state of the Lorenz attractor evolves according to the following expression:
\begin{subequations}
    \begin{align}
    \mathbf{x}_{t+1}&=\mathbf{F}(\mathbf{x}_t) \mathbf{x}_t + \mathbf{w}_t \in \mathbb{R}^3,\\
    \mathbf{F}(\mathbf{x}_t)&\triangleq\exp\left(\begin{bmatrix}
        -10 & 10 & 0 \\
        28 & -1 & \mathbf{x}_{t-1} \\
        0 & \mathbf{x}_t & -\frac{8}{3}
    \end{bmatrix} \Delta_t \right) \in \mathbb{R}^{3 \times 3}.
\end{align}
\end{subequations}
We have generated trajectories using the $5$-th order ($J=5$) Taylor series expansion of $\mathbf{F}(\mathbf{x}_t)$ for $\Delta_t=0.02$, setting the state process noise in~\eqref{eq:DS_def_basic} to $\mathbf{Q}=0.1 \mathbf{I}_3$. An example simulated trajectory is provided in Fig. \ref{fig:lorenzExample}. To investigate our filter's ability to generalize to incorrect model knowledge, we trained two variants: one assuming $J=5$ (correct dynamics) and one assuming $J=2$ (incorrect dynamics). A system with $N=3$ sensor nodes was considered, with the graph connection probability set to $p_c=0.4$. The first node received a noisy estimate of the first two state coordinates, the second observed the last two, and the third node receives an estimate of the first and third state coordinates after they have been converted to polar coordinates. The observation noise was additive zero-mean Gaussian with covariance $\mathbf{R}_i=\sigma_{\text{Lor}} \mathbf{I}_2$ for each $i$-th node. To ensure the neural filter generalizes across different temporal lengths rather than overfitting to a fixed sequence, the horizon $T$ for each trajectory was selected uniformly at random from the set\footnote{During training, trajectories were split to shorter segments of length $T_{\rm trunc}=20$ to apply truncated BPTT.} $\{1250, 1500, 1750, 2000\}$. 

Figure~\ref{fig:lorenzResults} depicts the average MSE performance with the observation noise strength $\sigma_{\text{Lor}}$ ranging from $-30$~dB to $0$~dB. The DPF benchmark displayed poor performance with very large errors, and has been omitted from the plot to maintain presentation quality. As shown in the figure, CA-NKCF achieves a significant performance gain, yielding up to $50 \%$ error reduction relative to the MF GRU benchmark, even when the former assumes incorrect system dynamics.

\subsection*{Sensitivity Study}
\begin{figure}[!t]
    \centering
    \includegraphics[width=\linewidth]{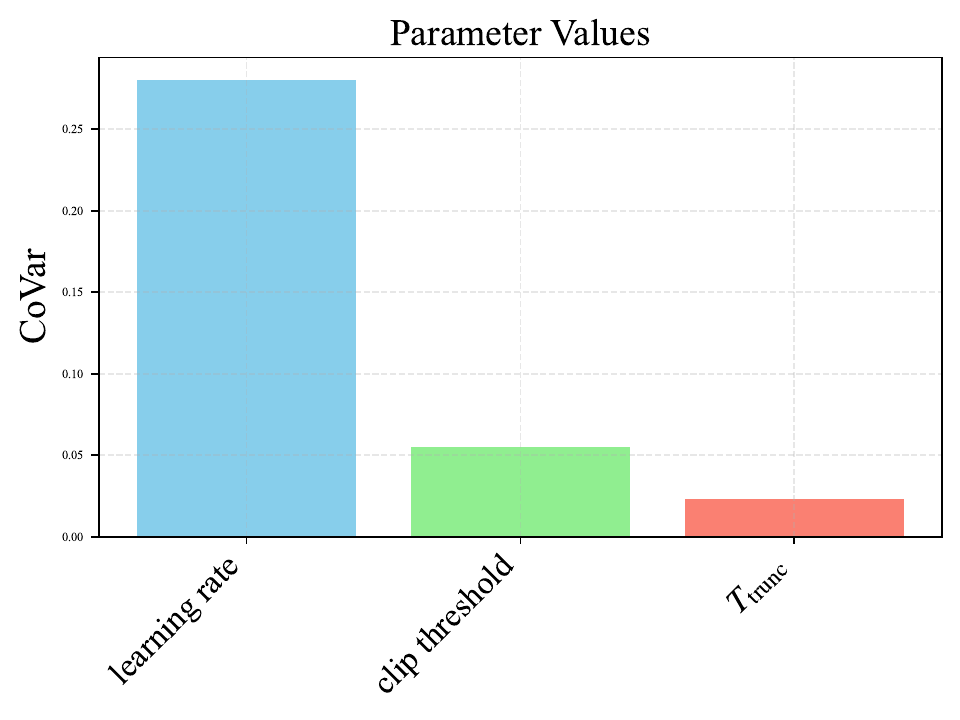}
    \caption{Sensitivity study results for the Lorenz attractor scenario considering observation noise of $0$~dB.}
    \label{fig:LorenzSensitivty}
\end{figure}

Training machine learning models on sequential data with a nonstationary structure can be an unstable process, highly affected by  hyperparameter choices~\cite{LSTMOdyssey}. Having established the effectiveness of the proposed CA-NKCF framework in the Lorenz attractor scenario, we have also used those chaotic trajectories to verify its stability to moderate hyperparameter changes. More specifically, we have tested the following: \begin{itemize}
    \item \textbf{Learning rate:} The training and testing procedures were repeated for learning rates ranging from $5\times10^{-5}$ to $10^{-3}$. 
    \item \textbf{Gradient clipping:} The clip threshold was modified from $0.5$ to $2.5$.
    \item \textbf{Truncation length ($T_{\rm trunc}$):} It was varied from $15$ to $50$.
\end{itemize}
The Coefficients of Variation (CoVar) on the test set, considering observation noise of $\sigma_{\text{Lor}}=0$ dB, are depicted in Fig.~\ref{fig:LorenzSensitivty}. As is evident, the proposed algorithm is robust to moderate hyperparameter changes as the CoVars are substantially smaller than $1$; this implies that the variance of the test set score is much smaller than the average. The learning rate causes some fluctuation in the final result, particularly for large values, whereas the other two parameters have negligible effect. 

\color{black}
\subsection{Wireless Tracking Application}
To demonstrate the applicability of the proposed estimator in real-world tasks, we have designed a practical wireless application inspired by the current \nth{5} Generation (5G) New Radio (NR) telecommunications standard~\cite{3gpp.38.211}. A wireless User Equipment (UE) tracking system~\cite{wymeersch2019radio} was considered, where the UE can move along two axes with a fixed acceleration  $(\alpha_x,\alpha_y)$. The UE state is the $4$-dimensional vector $\mathbf{x}_t=[p_{x,t},p_{y,t},v_{x,t},v_{y,t}]$, whose entries are dictated by: \begin{subequations}
    \begin{align}
    p_{x,t+1} &= p_{x,t} + v_{x,t}\Delta_t + \frac{1}{2}\alpha_x \Delta_t^2 + w_{p_x,t}, \\
    v_{x,t+1} &= v_{x,t} + \alpha_x \Delta_t + w_{v_x,t},  \\
    p_{y,t+1} &= p_{y,t} + v_{y,t}\Delta_t + \frac{1}{2}\alpha_y \Delta_t^2 + w_{p_y,t}, \\
    v_{y,t+1} &= v_{y,t} + \alpha_y \Delta_t + w_{v_y,t}.
    \label{eq:target_dynamics_explicit}
\end{align}
\end{subequations}
All state noise variables (the rightmost $w$ terms) were assumed zero-mean Gaussian with variance of $0.1$, the time interval was set to $\Delta_t=0.05$ sec, and the acceleration duplet was set to $(0.5,0.5)$ m/sec$^2$. 
The single-antenna UE was assumed to transmit a beacon signal at every time instance. Each sensor node corresponded to a BS equipped with $N_{\rm A}=8$ antenna elements, which was tasked to estimate the channel coefficients, i.e., the transfer function of the signal propagation system between itself and the UE. For each $i$-th BS, this channel coefficient is represented by $\mathbf{g}_{i,t}(\mathbf{p}_t)$, where $\mathbf{p}_t \triangleq [p_{x,t}, p_{y,t}]^\top$ is the current UE 2D coordinate vector, which can be estimated\footnote{Notice that the channel measurements are only determined by the current UE position vector. To this end, the  filter needs to incorporate past measurements through its GRU memory to estimate the velocity.} using existing channel estimation methods~\cite{italiano2024tutorial5gpositioning}. In our simulations, we have assumed the presence of noise during the channel estimation process, with a Signal-to-Noise Ratio (SNR) of $20$~dB, leading to erroneous observation of $\mathbf{g}_{i,t}(\mathbf{p}_t)$ denoted by $\hat{\mathbf{g}}_{i,t}(\mathbf{p}_t)$.
Besides the UE and the BSs, static objects known as scatterers were considered present in the scene. Those are, in general, passive environmental features (e.g., buildings, bridges, trees) that reflect the signal, creating multi-path conditions that adversely affect the estimation of the UE state. Note that, to ensure resolvability of the UE position from the considered multi-BS system, the total number of reception antennas must exceed the number of scatterers and, additionally, those antennas need to be spatially distributed, motivating the adoption of KCF-based techniques. Details of the channel model simulated are provided in the Appendix.

\begin{figure}
    \centering
    \includegraphics[width=\linewidth]{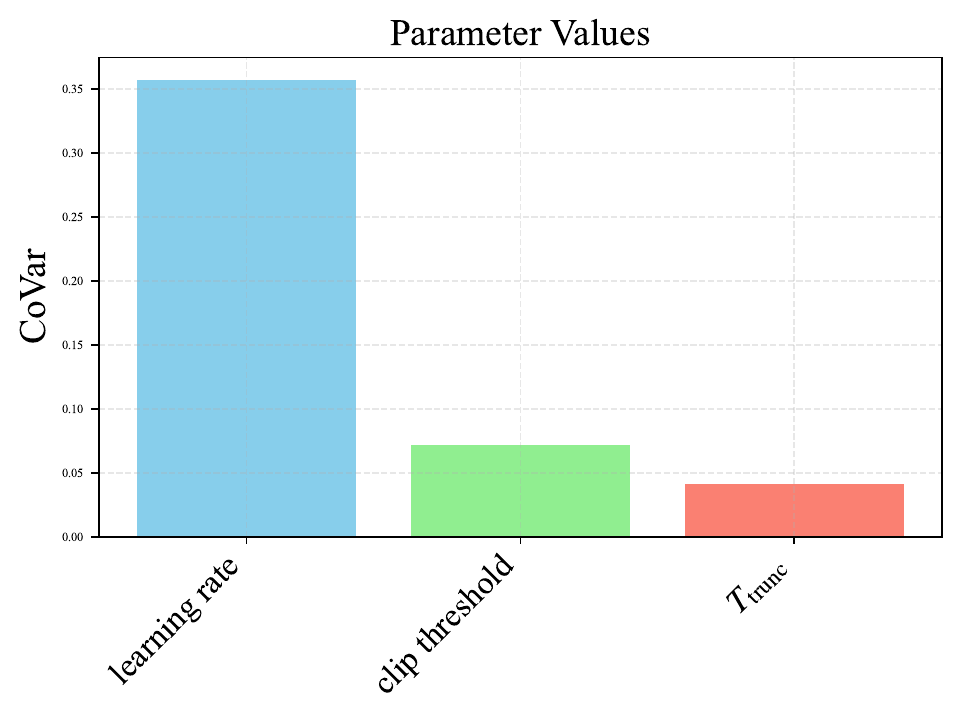}
    \caption{Sensitivity study results for the wireless UE tracking scenario considering $40$ scatterers.}
    \label{fig:wireless_sens}
\end{figure}

In Fig.~\ref{fig:WirelessResults}, we have fixed $T=50$, $p_c=0.4$, and varied the number of scatterers from $K=20$ to $50$ to evaluate the performance of the proposed CA-NKCF in rich scattering conditions. It is evident that the proposed UE position estimator outperforms all baselines by a pronounced margin. In addition, consistent with previous observations, the DPF benchmark exhibits particularly weak performance. While the UKCF serves as the most competitive baseline, CA-NKCF consistently surpasses it. Notably, even though our model's performance exhibits slight variance depending on the random initialization seed, its worst-case execution strictly outperforms the UKCF across all evaluated scattering densities. This experiment verifies that our consensus-based distributed estimation framework can achieve good performance in real-world systems.
Finally, after repeating the sensitivity study of Fig.~\ref{fig:LorenzSensitivty}
for this wireless tracking application, we verified the stability of the our MA algorithm in practical applications; the results for $K=40$ scatterers are depicted in Fig.~\ref{fig:wireless_sens}.

\begin{figure}[!t]
    \centering
    \includegraphics[width=\linewidth]{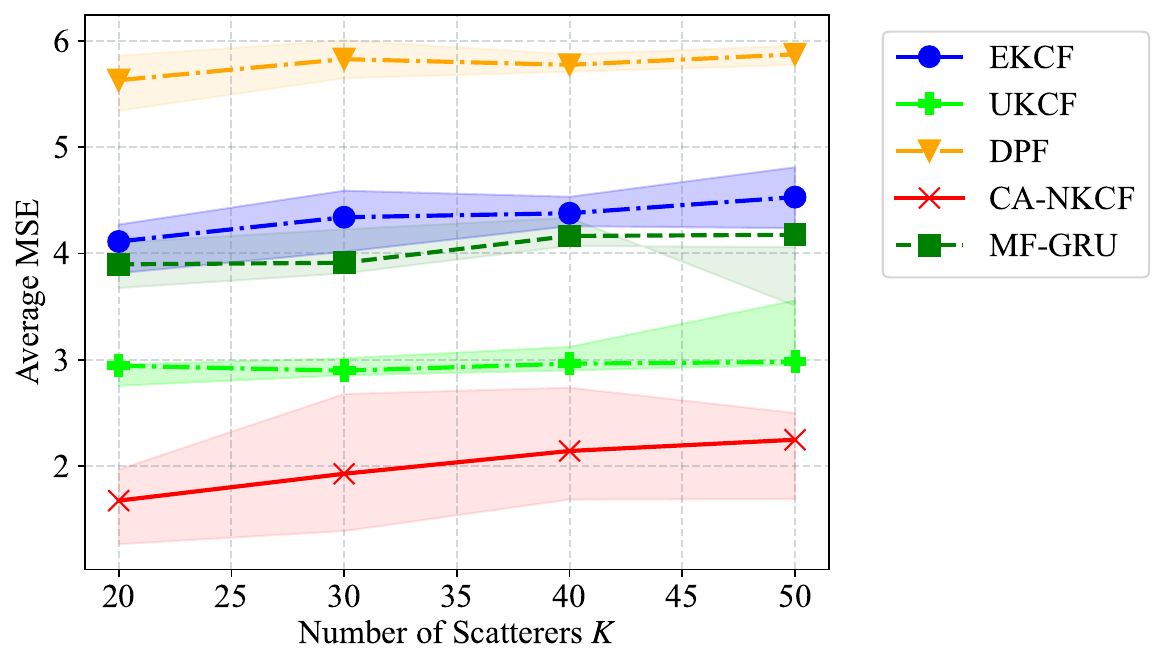}
    \caption{Average MSE versus the number of scatterers for the wireless UE tracking scenario.}
    \label{fig:WirelessResults}
\end{figure}
\subsection{Forward Inference Time Comparison}

While the preceding evaluations demonstrate the superior estimation capabilities of our hybrid scheme, that combines Kalman-like priors with data-driven NNs, compared to traditional latent state estimation algorithms, it is crucial to verify that the integration of GRUs in CA-NKCF does not incur additional latency. Table \ref{tab:runTime} details the forward inference times of all simulated estimation methods, confirming that our approach avoids computational bottlenecks. All evaluations were conducted on a $64$-bit Linux workstation. The system is equipped with an $11$th Generation Intel Core i$7-11700$KF processor ($8$ cores, $16$ threads) operating at a base frequency of $3.60$~GHz, alongside an NVIDIA GeForce RTX $3080$ GPU and $32$~GB of system memory. As shown from this table, the proposed CA-NKCF achieves faster execution times than standard MB estimators; this is primarily attributed to the fact that it bypasses the need for costly matrix inversions. Crucially, this efficiency ensures CA-NKCF satisfies the strict real-time causality constraint required for online tracking, as its inference time remains significantly smaller than the system's sampling period. In contrast, scaling particle-based baselines pushes their execution time beyond the physical state update interval, rendering their predictions obsolete.


\begin{table}[!t]
    \centering
    \begin{tabular}{|c|c|c|c|c|c|}
    \hline
         \textbf{Task}&\textbf{KCF}&\textbf{EKCF}&\textbf{UCKF}&\textbf{PF}&\textbf{CA-NKCF}  \\
         \hline
         \hline
         Linear $N=4$ & 0.0002&-&- &- &0.0001 \\
         \hline
         Linear $N=32$ & 0.0057& -&&- &0.0021 \\
         \hline
         Lorenz &-& 0.0002&0.0008 & 0.0102 &0.0001 \\
         \hline
         Wireless &- &0.0017&0.0003 &0.0024 & 0.0004\\
         \hline
    \end{tabular}
    \caption{Comparison of forward inference time in seconds between the proposed CA-NKCF and conventional filters.}
    \label{tab:runTime}
\end{table}

\section{Conclusion and Future Work}
In this paper, we presented CA-NKCF, a powerful domain-informed deep learning algorithm for online latent state estimation in decentralized systems. The proposed estimation framework combines the representational power of NNs, the temporal modeling abilities of GRUs, and the elegant mathematical structure of KFs with an optimized novel lightweight consensus mechanism. Strong mathematical intuition demonstrating that the learned consensus updates promote stable estimation was provided. Crucially, the proposed framework improves upon the communication costs of traditional information-based filters by requiring only the exchange of state priors, while completely bypassing the need for computationally expensive matrix inversions.
Detailed numerical investigations on physics-inspired trajectories and on a realistic wireless system under extreme scattering conditions showcased the superiority of our approach over both traditional MB and MF recurrent estimators. It was also demonstrated that the proposed hybrid architecture, hybrid combining Kalman-like priors with data-driven NNs, remains remarkably robust to reasonable hyperparameter changes.

For future work, we plan on extending CA-NKCF to tracking the state of more complicated latent processes, e.g., high-order autoregressive models and switching Markov models. Furthermore, we intend to develop distributed filters robust to Byzantine attacks and freeloaders. Mechanisms that adaptively optimize the agent network topology to improve distributed inference, leveraging bandit algorithms or reinforcement learning, constitute another interesting research direction.

\appendix
\section{Wireless Channel Model}

We have considered a wireless system operating at the carrier frequency $f_0 = 28$~GHz, comprising a single-antenna UE broadcasting a beacon signal (i.e., fixed and known) at every time instance $t$, which is received by a set of $N$ BSs playing the role of distributed sensor nodes. Each BS was assumed equipped with a Uniform Linear Array (ULA) of $N_{\rm A}=8$ antenna elements. The ULAs of all BSs were aligned with the $y$ axis of the coordinate plane, with each BS's adjacent antenna elements spaced at distances of $\lambda/2$, where $\lambda = c/f_0$ represents the wavelength and $c=3\times 10^8$ m/sec is the speed of light. Since, at every time-slot, only a single transmission took place, we assumed the UE to remain quasi-static, therefore Doppler effects were ignored. The wireless environment additionally contained $K$ scattering objects that partially absorbed and reflected the impinging signals; the positions and properties of all scatterers remained fixed within each UE trajectory. As a result, the channel in the far field between the UE and each $i$-th BS ($i=1,2,\ldots,N$) can be expressed, similar to~\cite{sayeed2002deconstructing}, as follows:
\begin{equation}\label{eq:channel}
    \mathbf{g}_{i,t}(\mathbf{p}_t) \triangleq \mathbf{g}_{i,t}^{\rm D} + \sum_{k=1}^K \mathbf{g}_{i,k,t} \in \mathbb{C}^{N_{\rm A}\times1},
\end{equation}
where $\mathbf{g}_{i,t}^{\rm D}$ is the direct link from the UE to the $i$-th BS and $\mathbf{g}_{i,k,t}$ is the link corresponding to the path reflected by the $k$-th scatterer (modeled as a point). The
$\mathbf{g}_{i,t}^{\rm D}$ is defined as:
\begin{equation}\label{eq:direct-link}
    \mathbf{g}_{i,t}^{\rm D} \triangleq L(\mathbf{\mathbf{p}}_t, \mathbf{p}^{\rm BS}_i) \exp\left( -\jmath\frac{2\pi}{\lambda} \| \mathbf{p}_t -\mathbf{p}^{\rm BS}_{i}\|_2 \right) \mathbf{a}(\theta^{\rm D}_{i,t}),
\end{equation}
where $\mathbf{p}^{\rm BS}_i$ is the position of the $i$-th BS (corresponding to its left-most ULA antenna element), $L(\mathbf{\mathbf{p}_t}, \mathbf{p}^{\rm BS}_i)$ represents signal attenuation due to pathloss: 
\begin{equation}\label{eq:pathloss}
    L\left(\mathbf{\mathbf{p}_t}, \mathbf{p}^{\rm BS}_i\right) =\left(\frac{\lambda}{4\pi \left\| \mathbf{\mathbf{p}_t} -\mathbf{p}^{\rm BS}_i\right\|_2}\right)^{2},
\end{equation}
the exponential factor models the distance-dependent phase shift, and $\mathbf{a}(\theta^{\rm D}_{i,t})$ is the ULA steering vector which depends on the angle $\theta^{\rm D}_{i,t}$ between the UE  antenna and the first antenna element of the $i$-th BS. This vector models incremental phase shifts incurred by the antenna spacing, as follows:
\begin{equation}\label{eq:steering-vector}
    \mathbf{a}(\theta^{\rm D}_{i,t}) \triangleq \begin{bmatrix} 1, e^{-j \pi \sin(\theta^{\rm D}_{i,t})}, \dots, e^{-j \pi (N_{\rm A}-1) \sin(\theta^{\rm D}_{i,t})} \end{bmatrix}^T.
\end{equation}
Moreover, each vector $\mathbf{g}_{i,k,t}$ in~\eqref{eq:channel} can be expressed as:
\begin{align}\label{eq:reflected-link}
    &\mathbf{g}_{i,k,t} = \Gamma_k L(\mathbf{\mathbf{p}}_t, \mathbf{p}^{\rm SC}_k) L(\mathbf{p}^{\rm SC}_k, \mathbf{p}^{\rm BS}_i) \\
    & \times \exp\left( -\jmath \frac{2\pi}{\lambda} \left\| \mathbf{p}_t -\mathbf{p}^{\rm SC}_{k}\right\|_2 + \left\| \mathbf{p}^{\rm SC}_{k} -\mathbf{p}^{\rm BS}_{i}\right\|_2 \right) \mathbf{a}(\theta^{\rm SC}_{i,k}),\nonumber
\end{align}
where $\Gamma_k$ is the $k$-th scatterer reflection coefficient modeled as a complex random value with uniform amplitude and phase, $\mathbf{p}^{\rm SC}_{k}$ is its position, and $\theta^{\rm SC}_{i,k}$ denotes the angle between the $k$-th scatterer and the $i$-th BS.

To emulate the channel estimation process at the fixed SNR level of $20$ dB, we have inserted additive white Gaussian noise to the actual channel vector as follows:
\begin{equation}\label{eq:channel-estimation}
    \hat{\mathbf{g}}_{i,t}(\mathbf{p}_t) = \mathbf{g}_{i,t}(\mathbf{p}_t) + \mathbf{v}_{i,t},
\end{equation}
where $\mathbf{v}_{i,t}$ was sampled from the complex normal distribution $\mathcal{CN}(\mathbf{0}_{N_{\rm A}}, \sigma_{\rm CE}^2 \mathbf{I}_{N_{\rm A}})$ with $\sigma_{\rm CE} = \| \mathbf{g}_{i,t}(\mathbf{p}_t) \|_{\rm F}/(10\sqrt{N_{\rm A}})$.
Therefore, following~\eqref{eq:channel}, the observation vector at each $i$-th BS sensor can be expressed as follows:
\begin{equation}\label{eq:wireless-observation-vector}
    \mathbf{z}_{i,t} = [\mathfrak{Re}(\hat{\mathbf{g}}_{i,t}(\mathbf{p}_t)), \mathfrak{Im}(\hat{\mathbf{g}}_{i,t}(\mathbf{p}_t))]^T \in \mathbb{R}^{2 N_{\rm A}\times1},
\end{equation}
where $\mathfrak{Re}(\cdot)$ and $\mathfrak{Im}(\cdot)$ represent the real and imaginary components of the channel estimation vector $\hat{\mathbf{g}}_{i,t}(\mathbf{p}_t)$ which were appended together to form $\mathbf{z}_{i,t}$. For our simulations, we have used $N=3$ BSs whose positions $\mathbf{p}^{\rm BS}_i$'s remained in the fixed positions $(10,10)$ m,  $(90,10)$ m, $(50, 95)$ m across all UE trajectories. At the beginning of every trajectory, $\mathbf{p}^{\rm SC}_{k}$'s were re-sampled uniformly within the bounded 2D box indicated by the coordinates of the $N$ BSs.

\bibliographystyle{ieeetr}
\bibliography{references}

\end{document}